\documentclass{article}

\usepackage{microtype}
\usepackage{graphicx}
\usepackage{subfigure}
\usepackage{booktabs} 
\usepackage{amsmath,amssymb}
\usepackage{mathdots}
\usepackage{txfonts}

\usepackage[font=small,skip=0pt]{caption}
\usepackage{array,enumitem}
\usepackage{scrextend}

\usepackage{hyperref}

\usepackage[accepted]{icml2018}

\icmltitlerunning{Improved Training of GANs using Representative Features}

\begin{document}

\twocolumn[

\icmltitle{Improved Training of Generative Adversarial Networks \\ using Representative Features }




\begin{icmlauthorlist}
\icmlauthor{Duhyeon Bang}{to}
\icmlauthor{Hyunjung Shim}{to}
\end{icmlauthorlist}

\icmlaffiliation{to}{School of Integrated Technology, Yonsei University, South Korea}

\icmlcorrespondingauthor{Hyunjung Shim}{kateshim@yonsei.ac.kr}

\icmlkeywords{Machine Learning, ICML, Generative adversarial network, GAN training}

\vskip 0.3in

]



\printAffiliationsAndNotice{}  
\begin{abstract}
Despite the success of generative adversarial networks (GANs) for image generation, the trade-off between visual quality and image diversity remains a significant issue. This paper achieves both aims simultaneously by improving the stability of training GANs. The key idea of the proposed approach is to implicitly regularize the discriminator using representative features. Focusing on the fact that standard GAN minimizes reverse Kullback-Leibler (KL) divergence, we transfer the representative feature, which is extracted from the data distribution using a pre-trained autoencoder (AE), to the discriminator of standard GANs. Because the AE learns to minimize forward KL divergence, our GAN training with representative features is influenced by both reverse and forward KL divergence. Consequently, the proposed approach is verified to improve visual quality and diversity of state of the art GANs using extensive evaluations.
\end{abstract}
\section{Introduction}\label{Sec1}
Generative models aim to solve the problem of density estimation by learning the model distribution ${\mathit{P}}_{\mathit{model}}$, which approximates the true but unknown data distribution of ${\mathit{P}}_{\mathit{data}}$ using a set of training examples drawn from ${\mathit{P}}_{\mathit{data}}$ \cite{ref09}. The generative adversarial networks (GANs) \cite{ref10} family of generative models implicitly estimate a data distribution without requiring an analytic expression or variational bounds of ${\mathit{P}}_{\mathit{model}}$. GANs have been mainly used for image generation, with impressive results, producing sharp and realistic images of natural scenes. The flexibility of the model definition and high quality outcomes has seen GANs applied to many real-world applications, including super-resolution, colorization, face generation, image completion, etc. \cite{ref27,ref36,ref37,ref38}.

Training a GAN requires two separate networks with competitive goals: a discriminator, $D$, to distinguish between the real and fake data; and a generator, $G$, to create as real as possible data to fool the discriminator. Consequently, the generator implicitly models ${\mathit{P}}_{\mathit{model}}$, which approximates ${\mathit{P}}_{\mathit{data}}$. This problem may be formulated as a minimax game \cite{ref10}, 
\vskip -0.31in
\[ {\mathop{\mathrm{min}}_{G} \ }{\mathop{\mathrm{max}}_{D} } \mathrm{\ \mathop{\mathbb{E}}_{x\sim \mathrm{P}_{\mathrm{data}}}\left[\log(D(x))\right]} \ +\ \mathop{\mathbb{E}}_{z\sim \mathrm{P}_{\mathrm{z}}}\left[\log(1-\mathrm{D}\left(\mathrm{G}\left(\mathrm{z}\right)\right)\right]\ ,\]
\vskip -0.2in
where $\mathbb{E}$ denotes expectation, $x$ and $z$ are samples drawn from ${\mathit{P}}_{\mathit{data}}$ and ${\mathit{P}}_{\mathit{model}}$ respectively.

When the generator produces perfect samples (i.e., ${\mathit{P}}_{\mathit{model}} \equiv {\mathit{P}}_{\mathit{data}}$), the discriminator cannot distinguish between real and fake data, and the game ends because it reaches a Nash equilibrium.  

Although GANs have been successful in the image generation field, training process instabilities, such as extreme sensitivity of network structure and parameter tuning, are well-known disadvantages. Training instability produces two major problems: gradient vanishing and mode collapse. Gradient vanishing becomes a serious problem when any subset of ${\mathit{P}}_{\mathit{data}}$ and ${\mathit{P}}_{\mathit{model}}$ are disjointed such that the discriminator separates real and fake data perfectly; i.e., the generator no longer improves the data because the discriminator has reached its optimum \cite{ref07}. This produces poor results, because training stops even though ${\mathit{P}}_{\mathit{model}}$ has not learned ${\mathit{P}}_{\mathit{data}}$ properly. Mode collapse is where the generator repeatedly produces the same or similar output because ${\mathit{P}}_{\mathit{model}}$ only encapsulates the major or single modes of ${\mathit{P}}_{\mathit{data}}$  to easily fool the discriminator. 


The trade-off between image quality and mode collapse has been theoretically and empirically investigated in previous studies \cite{ref40,ref23}, and generally either visual quality or image diversity has been achieved, but not both simultaneously. Visual quality can be achieved by minimizing reverse Kullback-Leibler (KL) divergence, which is suggested in standard GANs including \cite{ref10}. Meanwhile, image diversity is strongly correlated with minimizing forward KL divergence \cite{ref07}. Recent techniques \cite{ref14, ref15, ref23} have introduced a gradient penalty to regularize the divergence (or distance) for training GANs, and break the trade-off. The gradient penalty smooths the learning curve, improving training stability. Consequently, the gradient penalty is effective to improve both visual quality and image diversity, and has been evaluated for various GAN architectures.

We propose an unsupervised approach to \emph{implicitly} regularize the discriminator using representative features. This approach is similar to the gradient penalty, in that it also aims to stabilize training and break the trade-off between visual quality and image diversity, but does not modify the GAN objective function (i.e., the same divergence or loss definition are employed as a baseline GAN). Rather, we introduce representative features from a pre-trained autoencoder (AE) and transfer them to a discriminator to train the GAN. 
Because the AE learns to minimize forward KL divergence, adding its representative features to the discriminator of standard GAN lead the discriminator to consider two divergences (i.e., reverse and forward KL). Since forward KL tends to average the overall modes of data distributions during training \cite{ref09}, our representation features provide the overall mode information. Meanwhile, the objective of baseline discriminator pursues the reserve KL, thus tends to choose a single (few) mode of the data distribution. 
In other words, the discriminator is implicitly interrupted by representative features for discrimination, and encouraged to consider the overall data distribution.


The pre-trained AE learns from ${\mathit{P}}_{\mathit{data}}$ samples and is then fixed. Isolating representative feature extraction from GAN training guarantees that the pre-trained AE embedding space and corresponding features have representative power.
Since the representative features are derived from the pre-trained network, they are more informative during early stage discriminator training, which accelerates early stage GAN training. In addition, representative features provide the overall mode information as mentioned earlier, thus preventing GANs from mode collapse.
Although the representative features no longer distinguish real and fake images in the second half of training, the discriminative features continue to learn toward improving the discrimination power. Note that the total loss of the proposed model consists of loss of representative and discriminative features, and the discriminator learns the balance between them from the training data automatically.  
Therefore, the proposed approach stably improve both visual quality and image diversity of generated samples. We call this new architecture a representative feature based generative adversarial network (\textbf{RFGAN}).


The major contributions of this paper are as follows.
\begin{enumerate}[topsep=0pt,itemsep=-1ex]
    \item We employ additional representative features extracted from a pre-trained AE to implicitly constrain discriminator updates. This can be interpreted as effectively balancing reverse and forward KL divergences, thus GAN training is stabilized. Consequently, we simultaneously achieve visual quality and image diversity in an unsupervised manner. 
    \item The proposed RFGAN framework can be simply extended to various GANs using different divergences or structures, and is also robust against parameter selections. The approach employs the same hyper-parameters suggested by a baseline GAN. 
    \item Extensive experimental evaluations show RFGAN effectiveness, improving existing GANs including those incorporating gradient penalty \cite{ref14,ref15,ref23}.
\end{enumerate} 
Section~\ref{Sec2} reviews recent studies and analyzes how the proposed RFGAN relates to them. Section~\ref{Sec3} discusses RFGAN architecture and distinctive characteristics, and Section~\ref{Sec4} summarizes the results of extensive experiments including simulated and real data. The quantitative and qualitative evaluations show that the proposed RFGAN simultaneously improved image quality and diversity. Finally, Section~\ref{Sec5} summarizes and concludes the paper, and discusses some future research directions.
\section{Related Work} \label{Sec2}
Various techniques have been proposed to improve GAN training stability, which mostly aim to resolve gradient vanishing and mode collapse. Previous studies can be categorized into two groups as discussed below. 

\begin{enumerate}
\item{GAN training by modifying the network design }

To avoid gradient vanishing, the minimax game based GAN formulation was modified to a non-saturating game \cite{ref10}, changing the generator objective function from $J(G) = {\mathbb{E}}_{z\sim P_{z}} \log( 1-D(G(z) ) )$ to $J(G) = -\frac{1}{2} {\mathbb{E}}_{z\sim P_{z}} \log( D(G(z)))$. This relatively simple modification effectively resolved the gradient vanishing problem, and several subsequent studies have confirmed this theoretically and empirically \cite{ref07,ref23}. 

\cite{ref06} first introduced GAN with a stable deep convolutional architecture (DCGAN), and their visual quality was quantitatively superior to a variant of GANs proposed later, according to \cite{ref30}. However, mode collapse was a major DCGAN weakness, and unrolled GANs were proposed to adjust the generator gradient update by introducing a surrogate objective function that simulated the discriminator response to generator changes \cite{ref11}. Consequently, unrolled GANs successfully solved model collapse. 

InfoGAN \cite{ref32} achieved unsupervised disentangled representation by minimizing the mutual information of auxiliary (i.e., matching semantic information) and adversarial loss. Additionally, \cite{ref12} proposed various methods to stabilize GAN training using semi-supervised learning and smoothed labeling.  

\item{Effects of various divergences.} 

In \cite{ref18}, the authors showed that the Jensen-Shannon divergence used in the original GAN formulation \cite{ref10} can be extended to different divergences, including f-divergence (f-GAN). KL divergence has been theoretically shown to be one of the causes of GAN training instability \cite{ref07, ref13}, and the Wasserstein distance was subsequently to measure the similarity between ${\mathit{P}}_{\mathit{model}}$ and ${\mathit{P}}_{\mathit{data}}$ to overcome this instability. Weights clipping was introduced into the discriminator to implement the Wasserstein distance (WGAN) \cite{ref13} to enforce the k-Lipschitz constraint. However, weight clipping often fails to capture higher moments of ${\mathit{P}}_{\mathit{data}}$ \cite{ref15}, and a gradient penalty was proposed to better model ${\mathit{P}}_{\mathit{data}}$. The discriminator becomes closer to the convex set using the gradient penalty as a regularization term \cite{ref14}, which effectively improved Gan training stability. Least squares GAN (LSGAN) replaces the Jenson-Shannon divergence, defined by the sigmoid cross-entropy loss term, with a least squares loss term \cite{ref35}, which can be essentially interpreted as minimizing Pearson $\tilde{\chi}^2$ divergence. 
\end{enumerate}

Most previous GAN approaches have investigated stable architecture or additional layers to stabilize discriminator updates, changing the divergence, or adding a regularization term to stabilize the discriminator. The proposed RFGAN approach can be classified into the first category, modifying GAN architecture, and is distinct from previous GANs in that features from the encoder layers of the pre-trained AE are transferred while training the discriminator. 

Several previous approaches have also considered AE or encoder architectures.
ALI \cite{ref41}, BiGAN \cite{ref17} and MDGAN \cite{ref16} proposed that ${\mathit{P}}_{\mathit{data}}$ samples should be mapped to the generator latent space generator using the encoder structure. This would force the generator latent space to learn the entire ${\mathit{P}}_{\mathit{data}}$ distribution, solving mode collapse. Although these are similar approaches, in that encoder layers are employed to develop the GAN, RFGAN uses an AE to provide a completely different method to extract representative features, and those features stabilize the discriminator. 

EBGAN \cite{ref39} and BEGAN \cite{ref40} proposed an energy based function to develop the discriminator, and such networks showed stable convergence and less sensitivity to parameter selection. They adopted AE architecture to define the energy based function, which served the discriminator. In contrast, RFGAN employs representative features from the encoder layers and retains conventional discriminator architecture to maintain its discriminative power. Another alternative approach extracted features from discriminator layers, applied a denoising AE, and used the output to regularize adversarial loss \cite{ref34}. The technique improved image generation quality, and a denoising AE was employed to ensure robust discriminator features. However, this differs somewhat from the proposed RFGAN approach, where the AE is used as the feature extractor. 

In contrast to previous approaches that trained the AE or encoder layers as part of the GAN architecture, RFGAN separately trains the AE to learn ${\mathit{P}}_{\mathit{data}}$ in an unsupervised manner. Thus, feedback is disconnected from ${\mathit{P}}_{\mathit{model}}$ when training the AE, and the focus is on learning the feature space to represent ${\mathit{P}}_{\mathit{data}}$. In addition, RFGAN does not utilize the decoder, avoiding problems such as image blur. 

\section{Representative Feature based GAN} \label{Sec3}

To resolve GAN training instability, we extract representative features from a pre-trained AE and transfer them to the discriminator; implicitly enforcing the discriminator to be updated by effectively considering both reverse and forward KL divergence. 

The aim of an AE is to learn a reduced representation of the given data, since it is formulated by reconstructing the input data after passing through the network. Consequently,feature spaces learnt by the AE are powerful representations to reconstruct the ${\mathit{P}}_{\mathit{data}}$ distribution.

Several studies have utilized AE functionality as a feature extractor for classification tasks through fine-tuning \cite{ref01,ref02}. However, a good reconstruction representation does not guarantee good classification \cite{ref03,ref04}, because reconstruction and discriminative model features are derived from different objectives, and hence should be applied to their appropriate tasks for optimal performance. 

When training a GAN, the discriminator operates as a binary classifier \cite{ref06}, so features extracted from the discriminator specialize in distinguishing whether the input is real or fake. Thus, discriminator features have totally different properties from AE features. Considering these different properties, we denote AE and discriminator features as representative and discriminative features, respectively. Although the original GAN formulation evaluates data generation quality purely based on discriminative features, we propose leveraging both representative and discriminative features to implicitly regularize the discriminator, and hence stabilize GAN training. 

This section describes the proposed RFGAN model, and the effects of the modified architecture for training the discriminator. We also investigate how this effect could overcome mode collapse and improve visual quality.

\subsection{ RFGAN Architecture }

\begin{figure}[t!]
  \vskip 0.1in
    \begin{center}
      \centerline{\includegraphics[width=0.97\columnwidth]{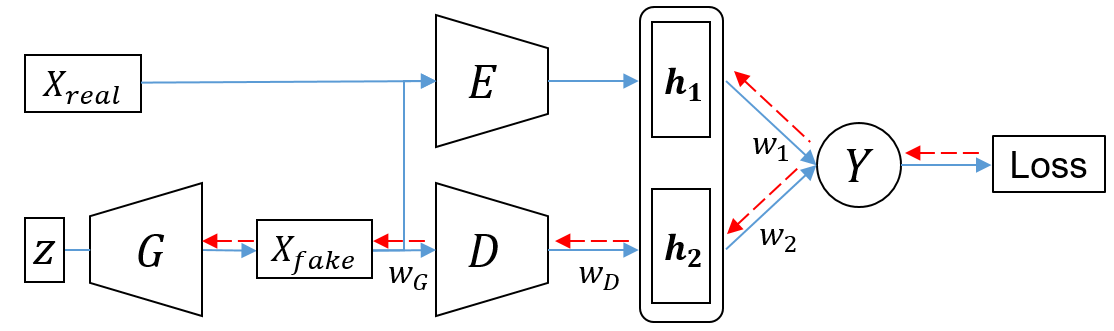}}
      \caption{Representative feature based generative adversarial network graphical model. $X_{real}$ and $X_{fake}$ are input and generated images, respectively; $E$, $G$, and $D$ are encoder, generator, and discriminator networks, respectively; $Z$ is the latent vector; $Y$ is binary output representing the real or synthesized image; $h_1$ and $h_2$ are representative and discriminative features, respectively; and $w_1$, $w_2$, $w_D$, and $w_G$ are network parameters. Blue solid and red dash lines represent forward and backward propagation, respectively.}
      \label{figure01}
    \end{center}
  \vskip -0.2in
\end{figure}

The main contribution of the RFGAN model is adopting representative features from a pre-trained AE to develop the GAN. Thus, RFGAN can be based on various GAN architectures, and refers to a set of GANs using representative features. For simplicity, we use DCGAN \cite{ref06} employing non-saturated loss as the baseline GAN, and apply representative features to the discriminator to construct DCGAN-RF. Section~\ref{Sec4} introduces various GANs as baselines to develop  RFGAN variants. We use exactly the same hyper-parameters, metrics, and settings throughout this paper, as suggested for a baseline GAN, to show that the RFGAN approach is insensitive to parameter selection, since representative features extracted are supplied from the encoder layer (part of the pre-trained AE) to the discriminator. The AE is pre-trained unsupervised using samples from ${\mathit{P}}_{\mathit{data}}$, and isolated from GAN training.

In particular, we construct the AE such that its encoder and decoder share the same architecture as the discriminator and generator, respectively. We then concatenate two feature vectors, one from the last convolution layer of the encoder and the other from the discriminator. Final weights are trained for the concatenated feature vector, to deciding between real or fake input. Figure~\ref{figure01} demonstrates the model for input data passing through encoder, $E$, and discriminator, $D$, networks; where$h_1$ and $h_2$ represent the representative and discriminative feature vectors, respectively, which are concatenated and transformed to a single sigmoid output, $Y$, through a fully connected layer. The output is evaluated with the ground truth label based on sigmoid cross entropy, and then the gradient of the loss function is delivered to the discriminator via backpropagation to update the parameters. This feedback is not propagated to the encoder, because its parameters are already trained and subsequently fixed. The procedure for gradient updates is 

$\ \ \mathit{D(x) = } - { \log{Y}}$ for ${x\sim \mathit{P}_{\mathit{data}}}$,  $Y=\sigmaup(h_1 w_1 + h_2 w_2)$,
\begin{flalign*}
    \mathit{\nabla }{\mathit{w}}_{\mathit{i}}=\frac{\partial D(x)}{\partial w_i}\mathit{=\ }-\frac{1}{Y}\cdot Y\left(1-Y\right)\cdot h_i=\left(Y-1\right)h_i, \mathit{i} \in \{1,2 \},
\end{flalign*}
\vskip -0.3in
\begin{align*}
    \mathit{\nabla }{\mathit{w}}_{\mathit{D}}=\frac{\partial D(x)}{\partial \mathit{w_D}}&=\left(Y-1\right)\cdot w_2 \cdot u\left(\mathit{w_D}\right).  \ \ \ \ \ \ \ \ \ \ \ \ \ \ \ \ \ \ \ \ \ \ \ \ \ \ \ \ \ \ \
\end{align*}
The GAN objective function represented by parameters
\begin{flalign*}
     \ \ \mathit{J}\left({\mathit{\thetaup }}_{\mathit{G}},{\theta }_D\right) &=\mathit{\mathop{\mathbb{E}}_{x\sim P_{data}}}\left[\log{D}\left(x;{\theta }\mathit{_D}\right)\right] \\
    &+\mathit{\mathop{\mathbb{E}}_{z\sim P_z}}\left[{\log \left(1-D\left(G\left(z;{\theta }_\mathit{G}\right);{\theta }_\mathit{D}\right)\right)\ }\right]&&
\end{flalign*}
is updated by
\begin{align*}
{\mathit{\thetaup }}^{\mathit{t+1}}_{\mathit{D}}\
    &\leftarrow \ {\mathit{\thetaup }}^{\mathit{t}}_{\mathit{D}}-\ {\eta }_\mathit{D}\frac{dJ\left({\theta }_\mathit{G},\ {\mathit{\thetaup }}^{\mathit{t}}_{\mathit{D}}\right)}{d{\mathit{\thetaup }}^{\mathit{t}}_{\mathit{D}}}=\ {\mathit{\thetaup }}^{\mathit{t}}_{\mathit{D}}-\ {\eta }_\mathit{D}\left({\mathit{\nabla }\mathit{w}}_{\mathit{D}}+\mathit{\nabla }{\mathit{w}}_{\mathit{i}}\right),\\
\mathit{\ \ \ }{\mathit{\thetaup }}^{\mathit{t+1}}_{\mathit{G}}\ 
    &\leftarrow {\mathit{\thetaup }}^{\mathit{t}}_{\mathit{G}}\ -\ {\eta }_\mathit{G}\frac{dJ\left({\theta }_\mathit{G},\ {\mathit{\thetaup }}^{\mathit{t}}_{\mathit{D}}\right)}{d{\mathit{\thetaup }}^{\mathit{t}}_{\mathit{G}}}\mathit{\ } .
\end{align*}
where $\mathit{\sigmaup }$ and $u$ are sigmoid and step functions, respectively.

Since the encoder is pre-determined, we only consider discriminator updates. We can derive the gradient toward the discriminator by calculating the partial derivative of loss term with respect to \textit{w${}_{D}$}, which indicates the network parameters as shown in Fig.~\ref{figure01}. Thus, $\mathit{\nabla }{\mathit{w}}_{\mathit{D}}$ depends on ${\mathit{h}}_{\mathit{1}}$, and the representative features affect the discriminator update. The procedure was derived for the case where $x$ is real. In the case of a fake sample, the same conclusion is reached, except that $D(x)$ is now $-{\log( 1-\mathit{\sigmaup }\mathit{(h_1w_1+h_2w_2)})\ }$.

Therefore, the generator is trained by considering both representative and discriminative features, because it should fool the discriminator by maximizing $\mathit{-}{\log \mathit{D(G(z))}\ }$. RFGAN representative features retain their properties, such as a global representation for reconstructing the data distribution, by fixing the encoder parameters.

\subsection{Mode collapse}

The AE decoder estimates the $P(x${\textbar}$En(x))$ distribution parameters based on a probabilistic interpretation, to generate $x$ with high probability formulated by cross-entropy loss \cite{ref08}.  It is possible to interpret that the AE follows forward KL divergence between ${\mathit{P}}_{\mathit{data}}$ and ${\mathit{P}}_{\mathit{model}}$ (i.e., KL(${\mathit{P}}_{\mathit{data}}$ {\textbar}{\textbar}${\mathit{P}}_{model}$)). Since the model approximated by forward KL divergence is evaluated using every true data sample (i.e., any $x : {\mathit{P}}_{\mathit{data}}(x) > 0$), it tends to average all ${\mathit{P}}_{\mathit{data}}$ modes \cite{ref09}. Hence, representative features extracted from the AE are similar, in that they effectively represent entire ${\mathit{P}}_{\mathit{data}}$ modes \cite{ref42}.

On the contrary, the aim of DCGAN with a non-saturated loss (the base architecture of the RFGAN model) is to optimize reverse KL divergence between ${\mathit{P}}_{\mathit{data}}$ and ${\mathit{P}}_{model}$, i.e., $\mathit{KL(}{\mathit{P}}_{\mathit{model}}\mathit{\ }\mathit{||}{\mathit{P}}_{\mathit{data}}\mathit{)\ -\ 2JSD}$ \cite{ref07}, where $JSD$ is Jensen--Shannon divergence. Since the reverse KL objective based model is examined for every fake sample (i.e., any $x : {\mathit{P}}_{\mathit{model}}\mathit{(x) > 0}$), it has no penalty for covering the entire true data distribution. Hence, it is likely to focus on single or partial modes of the true data distribution, which is the mode collapse problem.

The proposed RFGAN optimizes reverse KL divergence because the framework is built upon a non-saturated GAN. We also introduce AE representative features simultaneously, which encourages the model to cover the entire ${\mathit{P}}_{data}$ modes, similar to optimizing forward KL divergence. This suppresses the tendency toward mode collapse.

\subsection{Improving visual quality} 

\begin{figure}[t!]
  \vskip 0.1in
    \begin{center}
      \centerline{\includegraphics[width=\columnwidth]{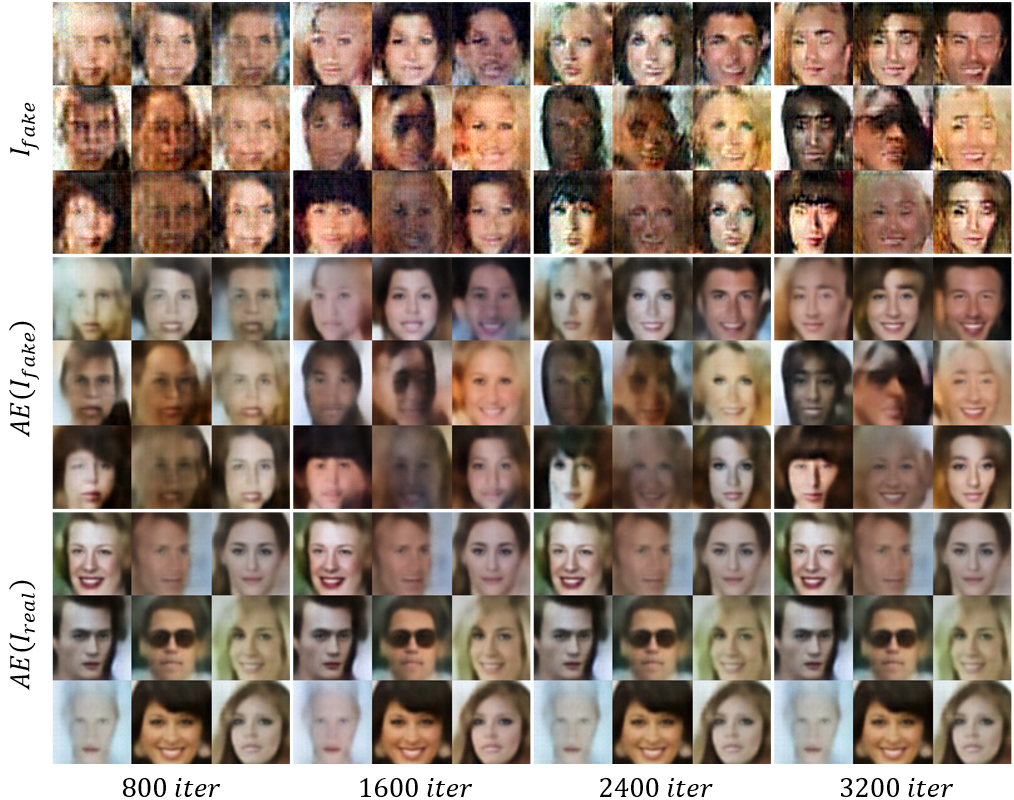}}
      \caption{Reconstruction comparison after iteration with the pre-trained AE. The first row shows generated images that are passed through the pre-trained AE along with real images, as shown in the second and third rows.}
      \label{figure02}
    \end{center}
  \vskip -0.4in
\end{figure}

Although representative features are useful to advance the discriminator in the early stage, they become less informative when approaching the second half of training, because the AE has limited performance for discrimination. Since the AE is built by minimizing reconstruction error, e.g. L2 or L1 loss; the model cannot learn multiple different correct answers, which causes the model to choose the average (or median) output \cite{ref09}. While this is useful to distinguish between poor fake and real input, when the generator starts producing good fake input, AE representative features are less discriminative, and hence interfere with decisions by the discriminator in the later training stages.  Figure~\ref{figure02} shows the output for several real and fake examples passed through the pre-trained AE. Real or fake inputs are easily distinguished at the beginning of training, but after several iterations they look similar. These experimental results demonstrate AE discriminative power for different levels of fake examples.

Thus, it is difficult to improve data generation visual quality beyond a certain level using representative features alone. Therefore, the proposed RFGAN model employs both representative and discriminative features to train the discriminator. Although representative features interfere with discrimination between real and fake input as training progresses, the RFGAN discriminator retains discriminative features, which allows training to continue. Consequently, the generator consistently receives sufficient feedback from the discriminator, i.e., the gradient from the discriminator increases, to learn ${\mathit{P}}_{\mathit{data}}$. Since these two features are opposing, and disagree with each other, the discriminator is stabilized without abnormal changes. 
By stabilizing discriminator growth, the RFGAN model generates high quality data, improving the original GAN.




\section{Experimental results} \label{Sec4}
For quantitative and qualitative evaluations, we include simulated and three real datasets: CelebA \cite{ref20}, LSUN-bedroom \cite{ref21}, and CIFAR-10 \cite{ref19}, normalizing between -1 and 1. A denoising AE \cite{ref31} is employed to improve feature extraction robustness, achieving a slight quality improvement compared to conventional AEs. 
Since the concurrent training of AE and GAN does not improve the performance, we use the pre-trained and then fixed AE for reducing computational complexity.

\subsection{Mode collapse }
\begin{figure}[t!]
  \vskip 0.2in
    \begin{center}
      \centerline{\includegraphics[width=\columnwidth]{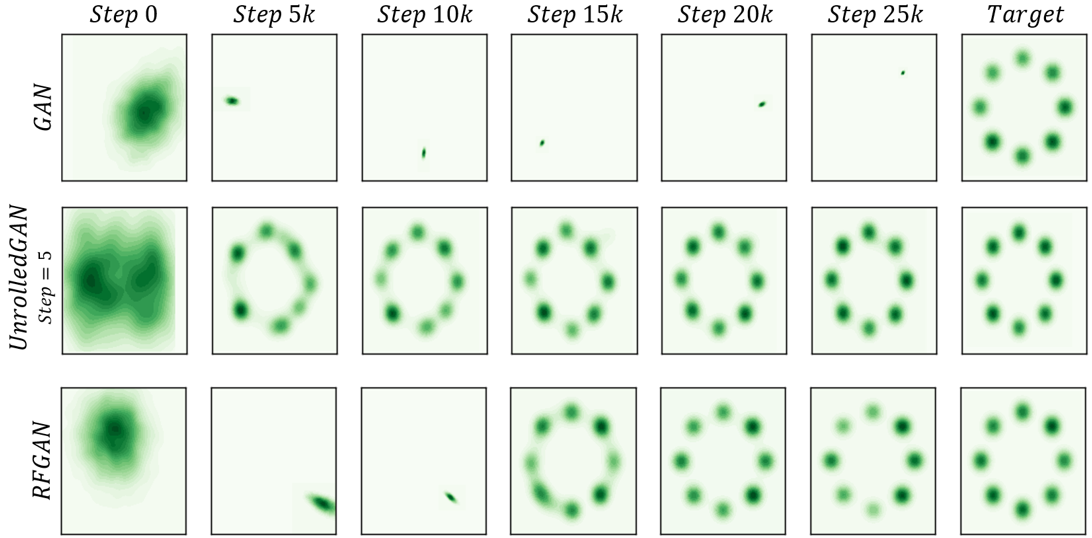}}
      \caption{Mode collapse test learning a mixture of eight Gaussian spreads in a circle}
      \label{figure03}
    \end{center}
  \vskip -0.3in
\end{figure}
To evaluate how well the RFGAN model could achieve data generation diversity, i.e., solving mode collapse, we train the network with a simple 2D mixture of 8 Gaussians \cite{ref11}. The Gaussian means form a ring, and each distribution has standard deviation = 0.1. Figure~\ref{figure03} compares RFGAN, GAN, and unrolled GAN models, and confirms that GAN suffers from mode collapse while unrolled GAN effectively solves this problem \cite{ref11}.

Previous studies solved mode collapse similarly to unrolled GAN by covering the entire distribution region and then gradually localizing the modes \cite{ref13,ref17}. However, RFGAN first learns each mode, and escapes from mode collapse by balancing representative features. This is because RFGAN minimizes reverse KL divergence, but is simultaneously influenced by representative features derived from forward KL divergence. When the representative features no long distinguish between real and fake input, the generator has achieved the representation power of the representative features. In other words, the generator learns the entire mode as well as the representative features, and then escapes mode collapse. Therefore, RFGAN first responds similarly to GAN, then gradually produces the entire mode. 

\subsection{Quantitative evaluation}
Since RFGAN is built upon the baseline architecture and its suggested hyper-parameters, input dimensionality is set at (64, 64, 3), which is acceptable for the CelebA and LSUN datasets. However, we modify network dimensions for the CIFAR-10 dataset, fitting the input into (32, 32, 3) to ensure fair and coherent comparison with previous studies. We also drew 500 k images randomly from the LSUN bedroom dataset for efficient training and comparison.

Two metrics were employed to measure visual quality and data generation diversity, respectively. The inception score \cite{ref12} measured visual quality for GANs using CIFAR-10 datasets, with larger score representing higher quality. The MS-SSIM metric is often employed to evaluate GAN diversity \cite{ref22}, with smaller MS-SSIM implying better the performance in producing diverse samples.

\begin{figure}[t!]
  \vskip 0.2in
  \centering
    \includegraphics[width=\columnwidth]{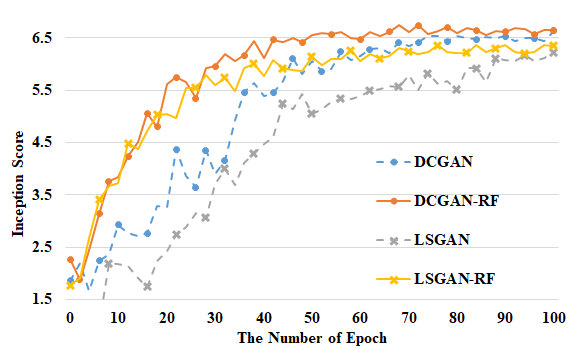}
    \vskip 0.2in
    {\Large
    \resizebox{0.8\columnwidth}{!}{\begin{tabular}{ccccc}
    \toprule
    & DCGAN & DCGAN-RF & LSGAN & LSGAN-RF \\
    \midrule
    Inception score & 6.5050 & 6.6349 & 5.9843 & 6.2791 \\
    \bottomrule
    \end{tabular}}
    }
    \caption{CIFAR10 inception score for DCGAN and LSGAN models with and without the representative feature approach}
    \label{figure04}
  \vskip -0.2in
\end{figure}

The inception score correlates well with human annotator quality evaluations \cite{ref12}, and hence is widely used to assess visual quality of GAN generated samples. We compute the inception score for 50 k GAN generated samples \cite{ref12}, using DCGAN based architecture to allow direct comparison with previous GANs. To show that the proposed algorithm was extendable to different GAN architectures, we also apply the proposed framework to other state of the art GANs (LSGAN \cite{ref35}, DRAGAN \cite{ref14}, and WGAN-GP \cite{ref15});  modifying their discriminators by adding representative features, and training them with their original hyper-parameters. The WGAN-GP generator is updated once after the discriminator is updated five times. 
Following the reference code\footnote{ https://github.com/carpedm20/DCGAN-tensorflow}, other networks are trained by updating the generator twice and the discriminator once. This ensures that discriminator loss did not vanish, i.e., the loss does not become zero, which generally provides better performance.

\begin{figure}[t!]
  \vskip 0.2in
  \centering
    \includegraphics[width=\columnwidth]{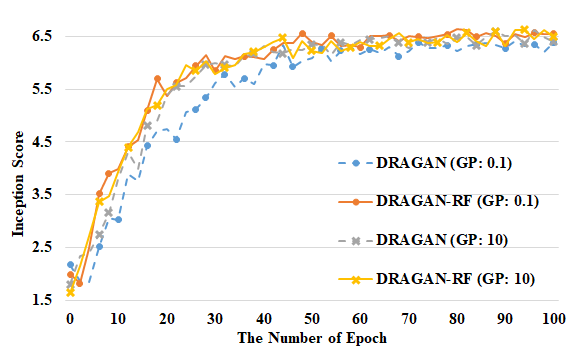}
    
    \includegraphics[width=\columnwidth]{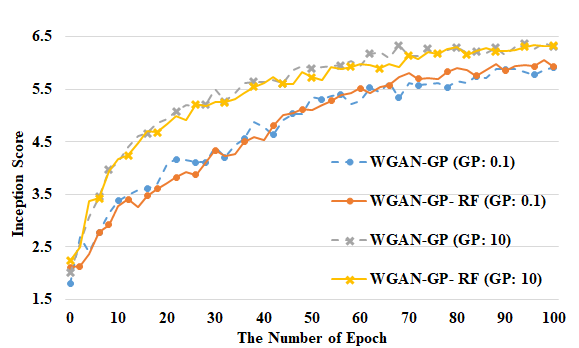}
    \vskip 0.2in
    {\Large
    \resizebox{0.9\columnwidth}{!}{\begin{tabular}{cccccc}
    \toprule
     \ & GP & DRAGAN & DRAGAN-RF & WGAN-GP & WGAN-GP-RF \\
    \midrule
    Inception & 0.1 & 6.3191 & 6.5314 & 5.7951 & 5.9141 \\
    Score & 10  & 6.4783 & 6.4905 & 6.2680 & 6.2699 \\
    \bottomrule
    \end{tabular}}
    }
    \caption{CIFAR10 inception score for (top) DRAGAN, and (bottom) WGAN-GP, with and without the representative feature for gradient penalty coefficients = 0.1 and 10}
    \label{figure05}
  \vskip -0.2in
\end{figure}

Figure~\ref{figure04} compares inception scores as a function of epoch. We compare DCGAN, DCGAN-RF, LSGAN, and LSGAN-RF, where ``-RF " extension refers to the base model with the representative feature. DCGAN-RF and LSGAN-RF outperform DCGAN and LSGAN, respectively, in terms of inception scores. In addition, DCGAN-RF and LSGAN-RF inception scores grow faster than DCGAN and LSGAN, respectively, confirming that the proposed representative feature improves training efficiency. Thus, the  proposed algorithm approaches the same visual quality faster than the baseline GAN.

The DRAGAN and WGAN-GP baseline GANs were recently proposed, using gradient penalty as a regularization term to train the discriminator. We also extend these using the proposed representative feature and compared with the original baseline GANs, as shown in Fig.~\ref{figure05}. The proposed modification still improves inception scores, although the improvement is not as significant as with DCGAN and LSGAN for coefficient of gradient penalty = 10. Interestingly, this coefficient plays an important role regarding the inception score, with larger coefficients producing stronger gradient penalty. Discriminator training is disturbed when the gradient penalty term became sufficiently strong, because gradient update is directly penalized. The score gap between DRAGAN and DRAGAN-RF increases when the coefficient of gradient penalty = 0.1, as expected since the DRAGAN performance approaches that of DCGAN as the gradient penalty decreases. However, since WGAN-GP replaces weight clipping in WGAN with a gradient penalty, it does not satisfy the k-Lipschitz constraint with low gradient penalty, which degrades WGAN-GP performance. Thus, it is difficult to confirm the tendency of the proposed representative feature for various WGAN-GP coefficients. 

The DCGAN-RF model produces the best overall score, including previous GANs with or without the proposed representative feature, which is consistent with previous studies that showed DCGAN to be the most effective model for high quality image generation \cite{ref30}. The proposed model achieves 0.128 mean improvement over the relevant baseline GAN, which is significant, and comparable or greater than the differences between different GANs. The improvement is particularly noticeable between LSGAN and LSGAN-RF.

\begin{table}[t!]
\caption{GAN diversity using the MS-SSIM metric. Real dataset MS-SIMM = 0.3727. NB: low MS-SSIM implies higher diversity.}
\label{table01}
  \vskip 0.1in
   \begin{center}
     \begin{Large}
      \begin{sc}
        \resizebox{0.8\columnwidth}{!}{\begin{tabular}{lcccc}
        \toprule
         & DCGAN & LSGAN & DRAGAN & WGAN-GP \\
        \midrule
        Original    & 0.4432 & 0.3907 & 0.3869 & 0.3813 \\
        with RF     & 0.4038 & 0.3770 & 0.3683 & 0.3773 \\
        \bottomrule
        \end{tabular}}
      \end{sc}
    \end{Large}
   \end{center}
  \vskip -0.2in
\end{table}

The MS-SSIM metric computes similarity between image pairs randomly drawn from generated images \cite{ref22}, and was  introduced it as a suitable measure for image generation diversity. However, MS-SSIM is meaningless if the dataset is already highly diverse \cite{ref23}. Therefore, we use only the CelebA dataset to compare MS-SSIM, since CIFAR-10 is composed of different classes, hence already includes highly diverse samples; and LSUN-bedroom also exhibits various views and structures, so has a diverse data distribution. We choose four previously proposed GANs as baseline algorithms: DCGAN, LSGAN, DRAGAN, and WGAN-GP, and compare them with their RFGAN variants (DCGAN-RF, LSGAN-RF, DRAGAN-RF, and WGAN-GP-RF, respectively), as shown in Table~\ref{table01}. the proposed GANs (RFGANs) significantly improve diversity (i.e., reduced MS-SSIM) compared with the baseline GANs, consistent over all cases, even in the presence of the gradient penalty term.

The LSGAN-RF, WGAN-GP-RF, and DRAGAN-RF scores are close to that of the real dataset diversity, i.e., the generator produces diverse samples reasonably well. DRAGAN-RF achieves the best MS-SSIM performance, generating the most diverse samples, whereas DCGAN-RF demonstrates the most notable improvement over the baseline (DCGAN), since DCGAN frequently suffers from mode collapse. Thus, the experimental study confirms that RFGAN effectively improved generated image diversity.

In addition to four baseline GANs, we also compare our results with ALI/BiGAN and AGE, which utilizes the encoders for GAN training. Since they focus on resolving mode collapse, the MS-SSIM is close to our results, but the inception score of ALI/BiGAN and AGE are much worse than our results; MS-SSIM of ALI/BiGAN and AGE is 3.7938 and 3.8133 respectively, and the inception score of ALI/BiGAN and AGE is and 5.34 and 5.90 respectively when our model achieves around 0.3816 of MS-SSIM and more than 6.20 of the inception score.

\subsection{Qualitative evaluation}
\begin{figure}[t!]
  \vskip 0.1in
    \begin{center}
    \centering
      \centerline{\includegraphics[width=1.05\columnwidth]{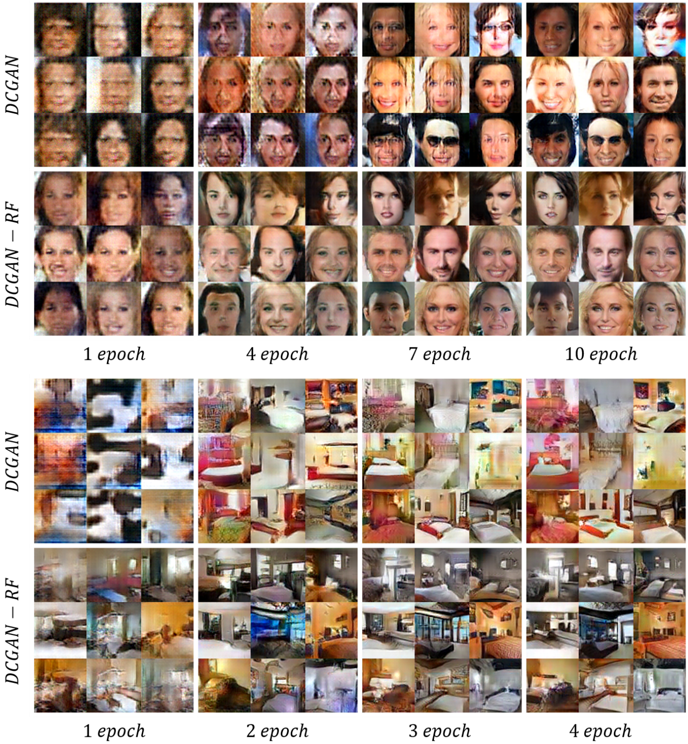}}
      \vskip -0.1in
      \caption{Stepwise visual quality comparison between generated images using DCGAN and DCGAN-RF trained with (left) CelebA and (right) LSUN}
      \label{figure06}
    \end{center}
  \vskip -0.3in
\end{figure}

\begin{figure}[t!]
  \vskip 0.1in
    \begin{center}
      \centerline{\includegraphics[width=1\columnwidth]{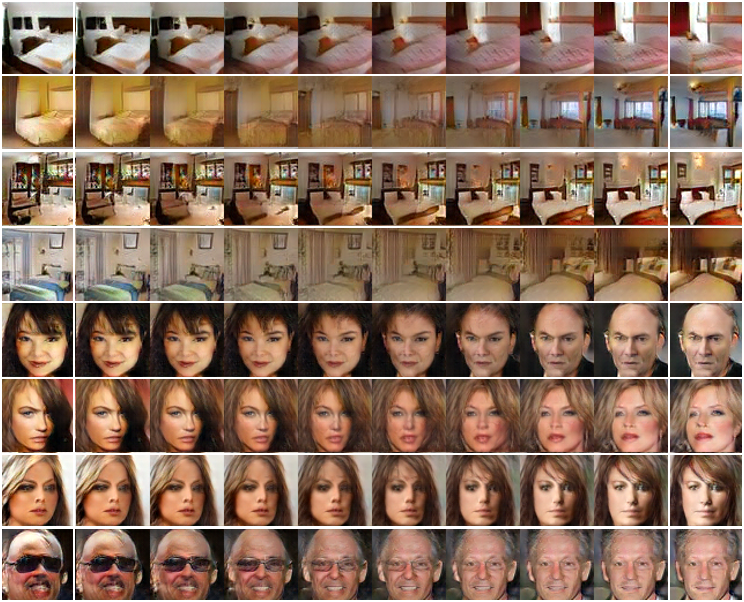}}
      \vskip -0.1in
      \caption{Latent space interpolations from LSUN and CelebA datasets. Left and right-most columns are samples randomly generated by DCGAN-RF, and intermediate columns are linear interpolations in the latent space between them.}
      \label{figure07}
    \end{center}
  \vskip -0.4in
\end{figure}
We compare DCGAN and DCGAN-RF generated images from the same training iteration, as shown in Fig.~\ref{figure06}. The proposed RFGAN produces significantly enhanced results,  and also speeds up the training process, with RFGAN visual quality being similar to results from later DCGAN iterations, which is consistent with Fig.~\ref{figure04}.
.

Since we reuse the training data to extract representative features, it is possible the performance enhancement came from overfitting the training data. To demonstrate that the enhancements are not the result of data overfitting, we generate samples by walking in latent space, as shown in Fig.~\ref{figure07}. Interpolated images between two images in latent space do not have meaningful connectivity, i.e., there is a lack of smooth transitions \cite{ref06,ref28,ref29}. This  confirms that RFGAN learns the meaningful landscape in latent space, because it produces natural interpolations of various examples. Thus, RFGAN does not overfit the training data.

\section{Conclusions} \label{Sec5}

This study proposes an improved technique for stabilizing GAN training and breaking the trade-off between visual quality and image diversity. Previous GANs explicitly add regularization terms, e.g. gradient penalty, to improve training stability, whereas the proposed RFGAN approach implicitly hinders fast discriminator update growth, thus achieving stable training. RFGAN employs representative features from an AE pre-trained with real data.
Our model achieves stabilizing and improving GAN training because RFGAN is influenced by two different characteristics of reverse and forward KL; learning the average mode and choosing a single mode.
Consequently, we successfully improve generated sample visual quality and solve mode collapse. We also show that the proposed RFGAN approach is easily extendable to various GAN architectures, and robust to parameter selection. 

In the future, our framework can be extended to various directions. For example, it is possible to utilize other types of features or more proper architectures, or training schemes that could further improve GAN performance. Specifically, replacing the convolution layer of the encoder to the residual block improves the visual quality; the inception score is increased from 6.64 to 6.73 for DCGAN-RF. The current study provides a basis for work employing various features or prior information to better design GAN discriminators. 

\section{Acknowledgement}

This research was supported by 
the MSIT(Ministry of Science and ICT), Korea, under the “ICT Consilience Creative Program” (IITP-2018-2017-0-01015) supervised by the IITP(Institute for Information $\&$ communications Technology  Promotion),
the MSIT(Ministry of Science and ICT), Korea, under the ITRC(Information Technology Research Center) support program(IITP-2018-2016-0-00288) supervised by the IITP(Institute for Information $\&$ communications Technology Promotion),
the Basic Science Research Program through the National Research Foundation of Korea (NRF) funded by the MSIP (NRF-2016R1A2B4016236),
and
ICT R$\&$D program of MSIP/IITP. [R7124-16-0004, Development of Intelligent Interaction Technology Based on Context Awareness and Human Intention Understanding]

\bibliographystyle{icml2018}

\end{document}